\documentclass[runningheads]{llncs}
\usepackage[T1]{fontenc}
\usepackage{graphicx}
\usepackage{lscape}        % For landscape layout
\usepackage{float}
\usepackage{adjustbox} % For fitting table within page width
\usepackage{booktabs} % For improved table appearance
\usepackage{longtable}
\usepackage{pdflscape} % For landscape tables
\usepackage{rotating} % For rotating tables horizontally
\usepackage{multirow}
\usepackage{tabularx}
\usepackage{amssymb} % For additional math symbols
\usepackage[utf8]{inputenc}
\usepackage{textcomp}
\DeclareUnicodeCharacter{2713}{\checkmark} % ✓
\DeclareUnicodeCharacter{2717}{\xmark}     % ✗ (define \xmark manually or use another symbol)
\usepackage{pifont}
\newcommand{\xmark}{\ding{55}} % Cross mark
\usepackage{caption} % Add this in the preamble if not already included
\usepackage{amsmath}
\usepackage{orcidlink}

\usepackage{tikz,xcolor,hyperref}

\begin{document}
%

% Make LNCS \orcidID use our circle
\renewcommand{\orcidID}[1]{\orcidcircle{#1}}
\title{Hidden Leaks in Time Series Forecasting: How Data Leakage Affects LSTM Evaluation Across Configurations and Validation Strategies} % Evaluating LSTM Configurations and Validation Techniques 
%Systematic Configuration of Validation Techniques for DL-Based Time Series Forecasting

%investigate the impact of configeration standardization on DL model performance

%The pre-processing steps could reduce variability in the dataset, potentially improving the model’s ability to detect leakage patterns 
%Configuring Validation Strategies for LSTM-Based Time Series forecasting

\titlerunning{Configuring Validation Strategies for LSTM-Based Time Series forecasting } %Impact of Data Leakage on LSTM Time Series Evaluation

% If the paper title is too long for the running head, you can set
% an abbreviated paper title here
%
\author{Salma Albelali\inst{1,2}\,\orcidlink{0000-0002-2819-5511} \and
Moataz Ahmed\inst{1,3}\,\orcidlink{0000-0003-0042-8819}}

\authorrunning{S. Albelali and M. Ahmed}

\institute{
King Fahd University of Petroleum \& Minerals, Department of Information and Computer Science, Dhahran, Saudi Arabia
\and
Imam Abdulrahman Bin Faisal University, Department of Computer Science, Dammam, Saudi Arabia\\
\email{salbelali@iau.edu.sa}
\and
SDAIA-KFUPM Joint Research Center for Artificial Intelligence, Dhahran, Saudi Arabia\\
\email{g201907430@kfupm.edu.sa, moataz@kfupm.edu.sa}
}

\maketitle              % typeset the header of the contribution
\begin{abstract}

Deep learning models, particularly Long Short-Term Memory (LSTM) networks, are widely used in time series forecasting due to their ability to capture complex temporal dependencies. However, evaluation integrity is often compromised by data leakage—a methodological flaw where input-output sequences are constructed prior to dataset partitioning, allowing future information to unintentionally influence training. This study investigates the impact of data leakage on performance, focusing on how validation design mediates leakage sensitivity. Three widely used validation techniques—2-way split, 3-way split, and 10-fold cross-validation—are evaluated under both leaky (pre-split sequence generation) and clean conditions, the latter mitigating leakage risk by enforcing temporal separation during data splitting prior to sequence construction.%The experimental framework examines the influence of configuration choices, including training epochs, early stopping, input window size, and lag step.

The effect of leakage is assessed using \textit{RMSE Gain}, which measures the relative increase in RMSE caused by leakage, calculated as the percentage difference between leaky and clean setups. Empirical results show that 10-fold cross-validation exhibits RMSE Gain values up to 20.5\% at extended lag steps. In contrast, 2-way and 3-way splits demonstrate greater robustness, typically maintaining RMSE Gain below 5\% across diverse configurations. Moreover, input window size and lag step significantly influence leakage sensitivity: smaller windows and longer lags increase the risk of leakage, whereas larger windows help reduce it. These findings underscore the need for configuration-aware, leakage-resistant evaluation pipelines to ensure reliable performance estimation.

%The abstract should briefly summarize the contents of the paper in 150--250 words.


\end{abstract}

\section{Introduction}
\label{sec:introduction}

Deep learning has significantly advanced time series forecasting by enabling models to learn complex temporal patterns from large volumes of sequential data. Among various architectures, Long Short-Term Memory (LSTM) networks have played a pivotal role in modeling time-dependent relationships due to their ability to mitigate vanishing gradient issues and retain long-range dependencies. While recent architectures such as Transformers have gained attention, LSTMs remain widely adopted and benchmarked in applied forecasting pipelines.

Accurate performance estimation is a critical concern in the verification and validation of deep learning models, especially in time series applications where the assumptions of traditional validation techniques are frequently violated due to temporal dependencies. Improper evaluation not only misrepresents a model’s generalization capacity but also compromises the trustworthiness of downstream deployment. Data leakage refers to a methodological flaw in machine learning pipelines where information from outside the training set (typically future observations) unintentionally influences model training, thereby violating the independence between training and test data \cite{kapoor2022leakage}. In time series forecasting, a common form of leakage arises when sequence windows are generated prior to dataset partitioning, allowing future values to be embedded in the training set through overlapping temporal context. This results in overly optimistic evaluation metrics that do not reflect true generalization capability. A particularly underexamined form of data leakage arises when input-output sequences are generated prior to dataset partitioning. This flawed pre-splitting design—often influenced by the chosen validation technique—can inadvertently allow future information to leak into the training set, thereby violating temporal causality and inflating reported performance.

The interaction between data leakage and validation design has received limited attention in time series machine learning. One of the central objectives of this study is to assess how different validation techniques respond to data leakage in time series forecasting. By systematically comparing 2-way, 3-way, and 10-fold cross-validation under both clean (post-split) and leaky (pre-split) configurations, we aim to identify which techniques exhibit the highest RMSE Gain due to improper sequence handling. To achieve this, we adopt a configuration-centric evaluation framework that treats the validation pipeline itself as a testable component within the forecasting system. Our experiments vary key modeling parameters—including training epochs, early stopping, input window size, and lag step—to evaluate their interaction with leakage under each validation technique. Rather than focusing on model internals, our goal is to assess the reproducibility and trustworthiness of the evaluation process. To assess leakage, \textit{RMSE Gain} is used, capturing the relative increase in error between clean and leaky setups.

Ensuring leakage-free evaluation is essential for the trustworthy deployment of forecasting models in real-world applications such as climate monitoring, energy demand prediction, and financial risk assessment---domains where misleading accuracy estimates can lead to costly or unsafe decisions. In this study, we use the Climate dataset and a univariate LSTM model to conduct six controlled experiments across varying validation techniques and temporal configurations. Our results indicate that 10-fold cross-validation is the most vulnerable to data leakage, with RMSE gain exceeding 20.5\% in certain scenarios. In contrast, 2-way and 3-way validation techniques exhibit more robust and consistent behavior, with the 2-way split demonstrating the least sensitivity to configuration changes.

\vspace{1em}
\noindent\textbf{Research Questions.} This study is guided by the following questions:
\begin{itemize}
    \item \textbf{RQ1:} How does the order of sequence generation (pre- vs.\ post-split) influence evaluation integrity in LSTM-based time series forecasting?
    \item \textbf{RQ2:} How do different validation techniques (2-way, 3-way, 10-fold CV) respond to data leakage under varying modeling conditions?
    \item \textbf{RQ3:} Which configuration parameters (e.g., window size, lag step, epochs) mitigate or exacerbate leakage-induced performance distortion?
\end{itemize}

\vspace{1em}
\noindent\textbf{Contributions.} This work makes the following contributions to the field of testing and validation in time series machine learning:
\begin{itemize}
    \item A systematic comparison of 2-way, 3-way, and 10-fold validation techniques under both clean and leaky configurations in LSTM-based time series.
    \item Assessment of the impact of data leakage through RMSE Gain, quantifying the relative change in error between clean and leaky configurations to reveal evaluation bias.
    \item A configuration-aware validation methodology that explores how sequence generation, training setup, and evaluation design interact with leakage sensitivity.
    \item Empirical evidence and guidelines for designing leakage-resilient evaluation pipelines, contributing to more trustworthy and reproducible testing practices in temporal machine learning.
\end{itemize}
%----------------------%

%---------------------------------------------%

\section{Literature Review}
\label{sec:literature_review}

LSTM networks are widely used in time series forecasting due to their ability to model long-range dependencies. However, their performance is highly sensitive to configuration parameters such as window size, lag step, batch size, and training duration. Several studies have focused on enhancing LSTM performance through architectural modifications or experimental tuning. Baghoussi et al. \cite{baghoussi2024corrector} proposed Corrector-LSTM with built-in data correction during training. Dhake et al. \cite{dhake2023algorithms} and Singh et al. \cite{singh2024ga} explored optimization strategies for LSTM hyperparameters, while Makinde \cite{makinde2024optimizing} compared different optimizers to improve convergence and accuracy. Sonani et al. \cite{sonani2025stock} further extended LSTM capabilities by integrating graph neural networks for enhanced pattern representation.

%...................%
\begin{table*}[htbp]
\renewcommand{\arraystretch}{1.5}  % <- Added line to increase row spacing
\centering
\caption{Comparison of LSTM-Based Time Series Forecasting Studies}
\label{tab:LSTM_comparison}
\resizebox{\textwidth}{!}{%
\begin{tabular}{clcl|cccc|cccccc}
\toprule
\textbf{Study} & \textbf{Dataset} & \textbf{Model} & \textbf{Batch} & \textbf{Time Scale} & \textbf{Window} & \textbf{Slide} & \textbf{Step} & 
\textbf{RMSE} & \textbf{MAE} & \textbf{MAPE} & \textbf{MSE} & \textbf{EDE} & \textbf{R2} \\
\midrule

\multirow{2}{*}{\cite{LSTM6}} & Total electron content (TEC) data & LSTM & 36 & Daily & 36 & - & - & 2.55 & 1.75 & 14 & - & - & 0.88 \\
  & air quality (AQ) data & LSTM & 36 & Daily & 3 & - & - & 14.34 & 12.1 & 12.55 & - & - & 0.83 \\

\cite{LSTM5} & Displacement monitoring dataset & LSTM & - & - & - & - & - & 0.00306 & - & - & - & - & - \\

\cite{LSTM_window1} & Volve oilfield in North Sea \cite{D1} & LSTM & - & Daily & 14 & \checkmark & - & 7.76 & - & 2.61\% & - & - & - \\

\multirow{2}{*}{\cite{LSTM_important}} & The Sales Dataset \cite{D8} & LSTM & 20 & Daily & 24 & \checkmark & 1,2,3,4 & 3.971 & 3.257 & 0.029 & - & - & - \\
 & The Sales Dataset \cite{D8} & ARIMA & 20 & Daily & 24 & \checkmark & 1,2,3,4 & 8.681 & 6.481 & 0.061 & - & - & - \\

\multirow{3}{*}{\cite{TS_Filling_CNN}} & Monitored sensors (Italy-SIM) & CNN\_LSTM & 512 & 15-min & 6 & \checkmark & 1 & - & 0.257 & 1.023 & 0.138 & - & 0.961 \\
 & Monitored sensors (Italy-SIM) & CNN-BiLSTM & 512 & 15-min & 6 & \checkmark & 1 & - & 0.176 & 0.709 & 0.068 & - & 0.981 \\
 & Monitored sensors (Italy-SIM) & LSTM & 512 & 15-min & 6 & \checkmark & 1 & - & 0.24 & 0.958 & 0.124 & - & 0.965 \\

\cite{LSTM_FillingData} & BD-IV plant stem moisture sensor & LSTM & 100 & 10 min & 200 & - & 1 & 0.847 & 0.626 & 1.001 & - & - & - \\

\multirow{5}{*}{\cite{LSTM_vessel}} & Aegean Sea Dataset & VLF-LSTM + Aug. & - & 30 min & 10 & \checkmark & 1 & - & - & - & - & 728 & - \\
 & Aegean Sea Dataset & SVM (RBF) & - & 30 min & 10 & \checkmark & 1 & - & - & - & - & 6744 & - \\
 & Aegean Sea Dataset & RFT & - & 30 min & 10 & \checkmark & 1 & - & - & - & - & 1903 & - \\
 & U.S. West Coast Dataset & VLF-LSTM + Aug. & - & 60 min & 10 & \checkmark & 1 & - & - & - & - & 2099 & - \\
 & Brest Dataset & VLF-LSTM + Aug. & - & 32 min & 10 & \checkmark & 1 & - & - & - & - & 2246 & - \\

\multirow{2}{*}{\cite{LSTM_validation}} & Huabei Basin & A-LSTM & 2 to 8 & Month & 6 & \checkmark & 1 & 0.0978 & 0.687 & 0.0167 & - & - & 0.9498 \\
 & Cambay Basin & A-LSTM & 2 to 8 & Month & 6 & \checkmark & 1 & 0.0102 & 0.0096 & 0.0096 & - & - & 0.9696 \\

\bottomrule
\end{tabular}}
% <- closes \resizebox
\end{table*}
%...................%

Table \ref{tab:LSTM_comparison} summarizes recent LSTM forecasting studies, highlighting common configuration elements such as time scale, window size, and prediction step. A few studies have attempted to improve temporal validation. Vamsikrishna and Gijo \cite{vamsikrishna2024new} proposed weighted k-fold methods to account for autocorrelation. Bergmeir et al. \cite{bergmeir2018note} emphasized the need to preserve temporal order in cross-validation, and Hewamalage et al. \cite{hewamalage2021recurrent} demonstrated that model performance is sensitive to data partitioning and preprocessing. Nevertheless, these works do not explicitly assess how validation techniques interact with sequence handling or model configuration under realistic time series conditions.

To address this gap, our study introduces a configuration-aware evaluation framework that treats the validation pipeline as a testable component. We systematically vary modeling parameters—window size, lag step, training duration, and early stopping—across both leaky (pre-split) and clean (post-split) setups. Using RMSE Gain as a diagnostic metric, we quantify the distortion introduced by data leakage and evaluate the sensitivity of 2-way, 3-way, and 10-fold cross-validation techniques. This approach provides empirical insights and practical guidelines for designing leakage-resilient evaluation procedures in time series forecasting.
%---------------------------------------------%

%-------------------------------------------------------%
\begin{table}[htbp]
\centering
\caption{Descriptive Statistics of the Dataset}
\label{tab:descriptive_stats1}
\scriptsize
\setlength{\tabcolsep}{4pt} % Reduce column separation
\renewcommand{\arraystretch}{1.1} % Adjust row spacing
\begin{tabular}{ll}
\toprule
\textbf{Metric} & \textbf{Value} \\
\midrule
Number of Samples & 1,462 \\
Time Range & 2013--2017 \\
\midrule
\multicolumn{2}{l}{\textbf{Target Variable: Mean Temperature (meantemp)}} \\
\midrule
Mean & 25.5 \\
Standard Deviation & 7.35 \\
Minimum & 6.00 \\
Median (50th Percentile) & 27.71 \\
Maximum & 38.71 \\
\bottomrule
\end{tabular}
\end{table}
%-----------------------------------------------------%

%%%%%%%%%%% D1_heatmap %%%%%%%%%%%%%%%%%%%
\begin{figure}[htbp]
  \centering
  \includegraphics[width=0.5\textwidth]{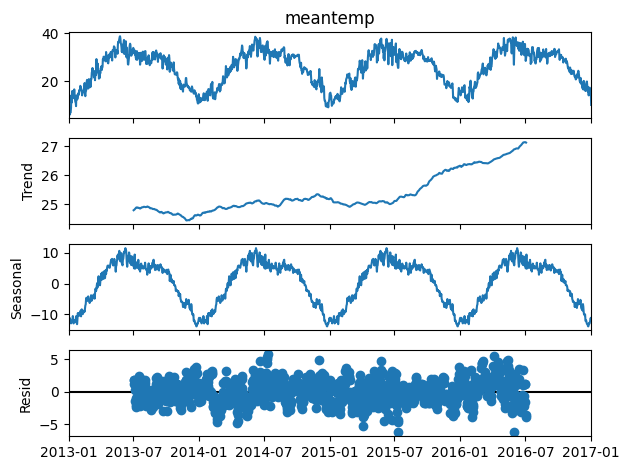}
  \caption{Seasonal Decomposition of Mean Temperature} %
  \label{Seasonal_Decomposition}
\end{figure}
%%%%%%%%%%%%%%%%%%%%%%%%%%%%%%%%%%%%%%

%---------------------------------------------%
\section{Materials and Methods}
\label{sec:methodology}

This study systematically evaluates the effect of experimental configurations on the stability and sensitivity of validation techniques in the presence of data leakage. To simulate this issue, a leaky evaluation pipeline is constructed by generating input-output sequences prior to dataset partitioning—an approach recognized as a source of temporal leakage and performance bias in time series modeling~\cite{hewamalage2021recurrent}. All baseline experiments, summarized in Table~\ref{experiment_setups_D1}, are initially conducted under this flawed pre-split setup to reflect common evaluation practices found in the literature. To assess how configuration settings influence leakage severity, several training parameters are varied, including the number of epochs, use of early stopping, and data splitting strategy (random vs.\ sequential). The analysis focuses on three commonly used validation techniques: 2-way holdout, 3-way split, and 10-fold cross-validation. In a second phase, the most leakage-sensitive configuration is re-evaluated under a clean setup in which sequence windows are generated only after data partitioning. This controlled comparison enables a quantifiable analysis of leakage-induced distortion. The evaluation is further extended by varying sequence modeling parameters related to data sufficiency, specifically input window size and lag step. This design enables assessment of how modeling choices interact with validation technique robustness. To measure the impact of leakage, RMSE Gain is used to show the relative increase in prediction error between clean and leaky configurations.

\subsection{Dataset and Forecasting Model}

\textit{Dataset description.} Experiments use the Daily Climate dataset comprising 1,462 daily weather records from January 2013 to April 2017. The univariate forecasting target is mean temperature. Descriptive statistics are reported in Table~\ref{tab:descriptive_stats1}, and the source is documented by Rao~\cite{D1_rao_2019}. Seasonal decomposition (Figure~\ref{Seasonal_Decomposition}) reveals strong annual cycles and a subtle warming trend, indicating the need for models that capture both seasonal and long-term dynamics.

\textit{Model configuration.} A univariate LSTM model is implemented in TensorFlow and Keras. The architecture includes a single LSTM layer with 64 units, followed by a dense output layer. The model is trained using the Adam optimizer (learning rate = 0.001), with mean squared error as the loss function.

\subsection{Validation Techniques}

Three validation strategies are evaluated:

\begin{itemize}
    \item \textbf{2-way split}: An 80\%/20\% sequential division into training and test sets.
    \item \textbf{3-way split}: A 70\%/10\%/20\% sequential split into training, validation, and test sets, supporting early stopping and tuning.
    \item \textbf{10-fold cross-validation}: Configurations use \texttt{KFold} with \texttt{shuffle=False} to preserve temporal ordering. In the leaky setup, sequence windows are generated before data splitting, allowing future information to bleed into training folds. In the clean setup, sequences are constructed post-split, ensuring strict chronological integrity and eliminating leakage.

\end{itemize}

This method enables a systematic assessment of how widely used validation paradigms respond to data leakage, providing insight into their robustness under realistic forecasting constraints.

\subsection{Experimental Design}
\begin{figure}[H]
  \centering
  \includegraphics[width=1.00\textwidth]{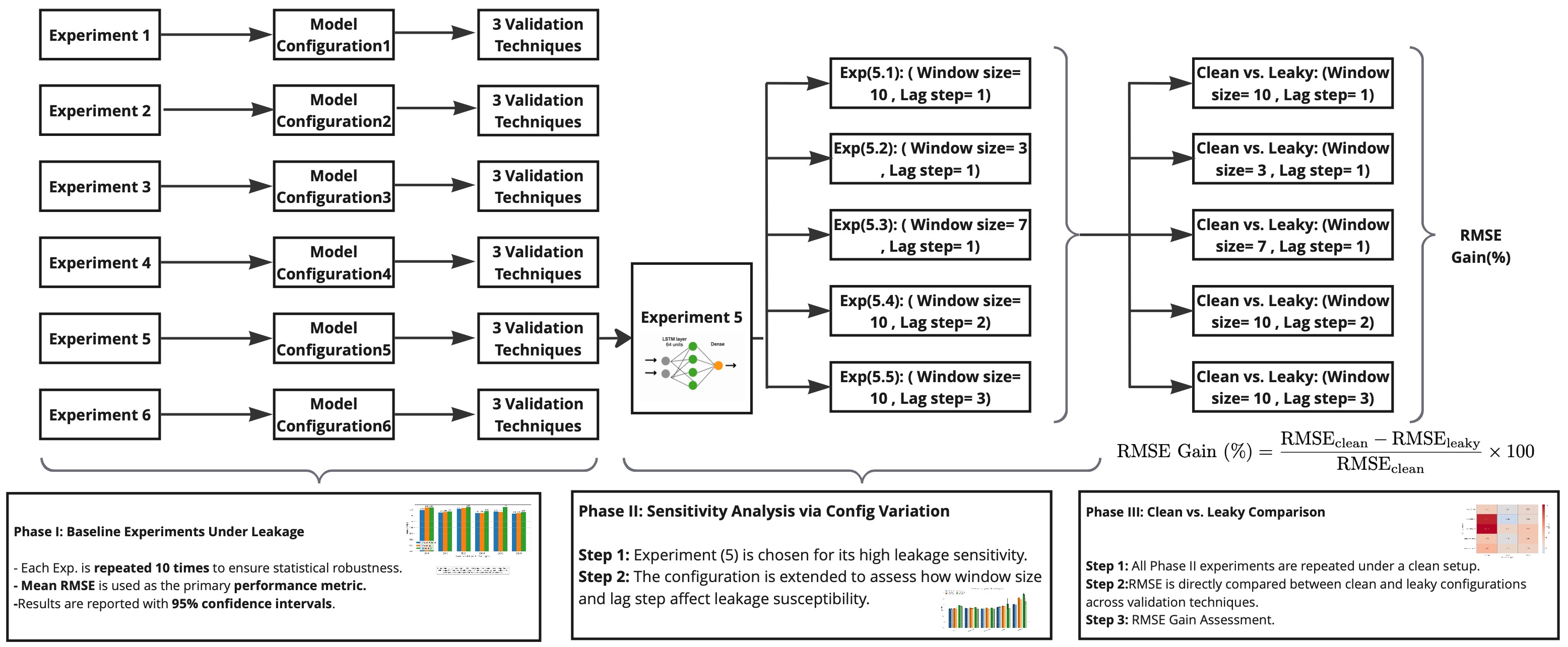}
  \caption{Methodology Framework.}
  \label{fig:methodology_diagram}
\end{figure}

The methodology follows a multi-phase evaluation pipeline, as illustrated in Figure~\ref{fig:methodology_diagram}. Each experiment is repeated 10 times for validation purposes without fixing random seeds to capture variability, resulting in a total of 480 experiments. The phases are as follows:

\subsubsection{Phase I: Baseline Experiments Under Leakage.}

In the first phase, six baseline experiments (Exp.~1–6) are conducted under the leaky configuration, where input-output sequences are generated prior to data splitting. Each experiment varies in training configuration, including the number of epochs (30, 50, 100, or 200), the use of early stopping (enabled or disabled with patience values of 5 or 10), and the data splitting strategy (random vs.\ sequential). For example, Exp.~1 uses 30 epochs with random splits, while Exp.~5 employs sequential splitting and early stopping (patience = 10) with blocked 10-fold cross-validation. Each setup is repeated 5–10 times to ensure statistical robustness, and performance is reported using root mean squared error (RMSE) along with 95\% confidence intervals. The detailed configurations and outcomes of these experiments are summarized in Table~\ref{experiment_setups_D1}. Results from this phase are used to identify the most leakage-sensitive configuration for further analysis in subsequent phases.

%%%%%%%%%%%%%%% Updated Table 1: experiment_results of D1 %%%%%%%%%%%%%%%%
\begin{landscape}
\begin{table}
\caption{Configuration Test: Experiment Setup and Results.}
\label{experiment_setups_D1}
\renewcommand{\arraystretch}{2.5} % Increase row height
\fontsize{10}{12}\selectfont % Set font size (10pt) and line spacing (12pt)
\resizebox{\linewidth}{!}{%
\begin{tabular}{ccccccccccccccc}
\toprule
\multicolumn{7}{c}{\textbf{Experimental Setup}} & \multicolumn{8}{c}{\textbf{Experimental Results}} \\
\cmidrule(lr){1-7} \cmidrule(lr){8-15}
\textbf{Exp.} & \textbf{Epochs} & \textbf{ES} & \textbf{Split} & \textbf{Validation Technique} & \textbf{Runs} & \textbf{Patience} & \textbf{Min of RMSE} & \textbf{Max RMSE} & \textbf{Standard Error} & \textbf{Mean RMSE} & \textbf{Confidence Interval} & \textbf{Optimal Epoch} & \textbf{Last Epoch} & \textbf{Standard Dev.} \\
\midrule
1 & 30 & - & Random & (80/20) 2-way split & 10 & - & 1.7 & 1.76 & 0.01 & 1.72 & (1.71, 1.73) & - & - & 0.02 \\
  &   &   &   & (70/10/20) 3-way split &   & - & 1.77 & 1.99 & 0.02 & 1.83 & (1.78, 1.87) & - & - & 0.06 \\
  &   &   &   & 10-fold cross validation &   & - & 1.81 & 1.85 & 0 & 1.83 & (1.82, 1.84) & - & - & 0.01 \\
\midrule
2 & 50 & - & Random & (80/20) 2-way split & 10 & - & 1.61 & 1.66 & 0.003 & 1.63 & (1.62, 1.63) & - & - & 0.01 \\
  &   &   &   & (70/10/20) 3-way split &   & - & 1.62 & 1.67 & 0 & 1.64 & (1.63, 1.65) & - & - & 0.01 \\
  &   &   &   & 10-fold cross validation &   & - & 1.66 & 1.69 & 0 & 1.67 & (1.67, 1.68) & - & - & 0.01 \\
\midrule
3 & 50 & - & Sequential & (80/20) 2-way split & 10 & - & 1.75 & 1.83 & 0.01 & 1.78 & (1.76, 1.79) & - & - & 0.03 \\
  &   &   &   & (70/10/20) 3-way split &   & - & 1.88 & 1.79 & 0.01 & 1.82 & (1.80, 1.84) & - & - & 0.03 \\
  &   &   &   & 10-fold cross validation &   & - & 1.84 & 1.87 & 0.01 & 1.85 & (1.84, 1.85) & - & - & 0.01 \\
\midrule
4 & 100 & \checkmark & Random & (80/20) 2-way split & 10 & 5 & 1.59 & 1.64 & 0.005 & 1.61 & (1.59, 1.62) & 100 & 100 & 0.017 \\
  &   &   &   & (70/10/20) 3-way split &   &   & 1.59 & 1.64 & 0.02 & 1.61 & (1.60, 1.62) & 67 & 100 & 0.02 \\
  &   &   &   & 10-fold cross validation &   &   & 1.66 & 1.70 & 0 & 1.68 & (1.67, 1.69) & 54.06 & 100 & 0.01 \\
\midrule
5 & 100 & \checkmark & Sequential & (80/20) 2-way split & 10 & 10 & 1.62 & 1.77 & 0.01 & 1.66 & (1.63, 1.69) & 44.8 & 100 & 0.04 \\
  &   &   &   & (70/10/20) 3-way split &   &   & 1.51 & 1.69 & 0.04 & 1.65 & (1.53, 1.76) & 54 & 100 & 0.05 \\
  &   &   &   & 10-fold cross validation &   &   & 1.82 & 1.89 & 0.01 & 1.86 & (1.84, 1.88) & 52.05 & 100 & 0.03 \\
\midrule
6 & 200 & \checkmark & Random & (80/20) 2-way split & 5 & 10 & 1.58 & 1.60 & 0 & 1.58 & (1.57, 1.59) & 194 & 200 & 0.01 \\
  &   &   &   & (70/10/20) 3-way split &   &   & 1.58 & 1.60 & 0 & 1.59 & (1.58, 1.60) & 83.6 & 200 & 0.01 \\
  &   &   &   & 10-fold cross validation &   &   & 1.63 & 1.65 & 0 & 1.64 & (1.63, 1.65) & 72.62 & 200 & 0.01 \\
\bottomrule
\end{tabular}%
}
\end{table}
\end{landscape}
%%%%%%%%%%%%%%%%%%%%%%%%%%%%%%%

%%%%%%%%%%% Methodology Overview %%%%%
%%%%%%%%%%%%%%%%%%%%%%%%%%%%%%%%%%%%%%

\subsubsection{Phase II: Sensitivity Analysis via Config Variation.}

Based on the results from Phase I, Experiment~5 is selected for further analysis due to its sensitivity to data leakage. In Phase II, this configuration is extended to evaluate the impact of two key sequence modeling parameters—input window size and lag step—on leakage susceptibility. Specifically, window sizes of 3, 7, and 10, and lag steps of 1, 2, and 3 are systematically tested to assess the adequacy of temporal context in forecasting. Each configuration is evaluated under the leaky setup across all three validation strategies to investigate how architectural adequacy parameters influence leakage susceptibility.

\subsubsection{Phase III: Clean vs. Leaky Comparison.}

To isolate the effect of data leakage, all Phase II experiments are replicated under a clean configuration, enabling a direct comparison of RMSE between leaky and leakage-free setups across different validation techniques, window sizes, and lag steps. The key distinction between the two configurations lies in the timing of sequence generation. In the \textit{leaky configuration}, input-output sequences are constructed from the full dataset prior to partitioning, allowing future information to unintentionally leak into the training process and compromise evaluation integrity. In contrast, the \textit{clean configuration} preserves temporal consistency by first splitting the dataset into training, validation, and test subsets, and then generating sequences independently within each partition, thereby eliminating the risk of information leakage. All experiments apply consistent preprocessing procedures and a fixed sequence length of 10, ensuring a fair and controlled basis for comparison between the two configurations.

%\subsection{Leakage Quantification via RMSE Drift}

This study assesses the impact of data leakage by measuring the effect on RMSE between clean and leaky experimental setups, expressed as \textit{RMSE Gain}:

\[
\text{RMSE Gain (\%)} = \frac{\text{RMSE}_{\text{clean}} - \text{RMSE}_{\text{leaky}}}{\text{RMSE}_{\text{clean}}} \times 100
\]

This metric captures the relative change in RMSE attributable to leakage and enables comparative evaluation of validation techniques and architectural configurations. A \textbf{positive RMSE Gain} indicates an apparent performance improvement under leaky conditions—i.e., the model’s error appears lower with leakage present (\(\text{RMSE}_{\text{leaky}} < \text{RMSE}_{\text{clean}}\)), reflecting an optimistic bias caused by leakage. Conversely, a \textbf{negative RMSE Gain} suggests performance degradation or stochastic variability, where the leaky setup results in higher error.

%--------------------------------------------%

%%%%%%%%%%%%%%%%%%%%%%%%%%%%%%%%%%%%%%%%%%%%%%%%%%%%%%%%%%%%%%%%

\section{Results and Discussion}
\label{sec:results}

Table~\ref{experiment_setups_D1} summarizes the RMSE outcomes for six baseline experiments (Exp.~1–6), each conducted under leaky conditions. The configurations systematically vary in terms of training epochs, early stopping criteria, and data splitting strategies (random vs.\ sequential). Across all configurations, the 2-way split consistently yielded the lowest mean RMSE values, especially under longer training configurations. For instance, Experiment~6 (200 epochs) achieved a mean RMSE of 1.58 with a standard deviation of 0.01, indicating strong predictive performance and minimal variability even in the presence of leakage.

These trends are further visualized in Figure~\ref{fig:baseline_rmse}, which compares average RMSE across validation techniques. The figure highlights that 10-fold cross-validation, particularly when applied with sequential data splitting, demonstrates a pronounced susceptibility to temporal leakage. Notably, Experiment~5 (100 epochs, sequential split) produced an RMSE of 1.86 under 10-fold CV, substantially higher than its 2-way and 3-way counterparts. This RMSE Gain can be attributed to repeated fold reuse across temporally adjacent windows, which increases the likelihood of information from future time steps leaking into training batches.

The baseline results confirm that both the structure of the validation technique and the configuration of training critically impact the model's vulnerability to leakage. Due to its elevated RMSE, wider confidence intervals, and greater sensitivity to sequence mismanagement, Experiment~5 was selected for in-depth analysis in Phases II and III.

%%%%%%%%

%\subsection*{3. Phase II — Sensitivity to Architectural Adequacy Parameters}

\textbf{Phase II:} Experiment 5 was extended to analyze how architectural parameters—window size and lag step—impact leakage sensitivity under leaky evaluation setups.

\begin{figure}[H]
  \centering
  \includegraphics[width=0.90\textwidth]{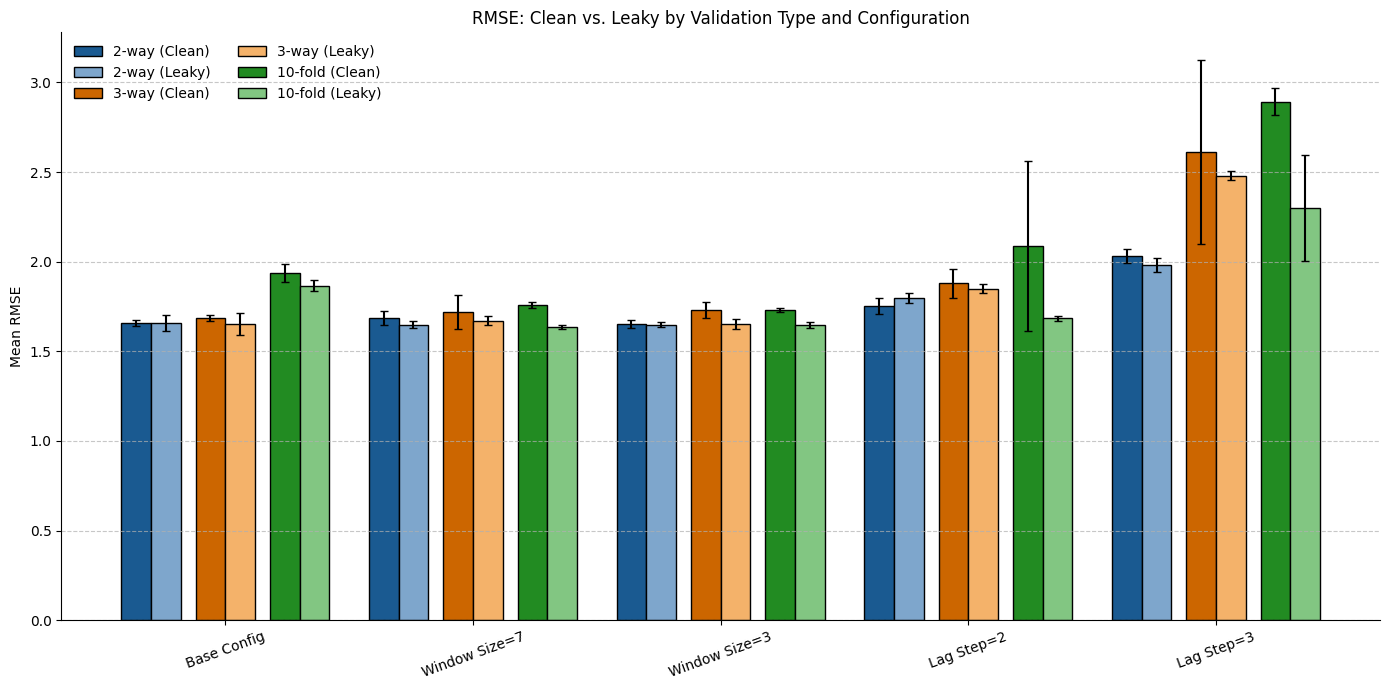}
  \caption{RMSE of Clean vs. Leaky Setups Across Validation Techniques. }
  \label{Sequence}
\end{figure}

\begin{itemize}
\item \textbf{Lag Step:} Increasing the lag step generally led to higher RMSE values and amplified leakage effects. Under 10-fold cross-validation, the RMSE decreased from 2.0868 (clean) to 1.6841 (leaky) at lag = 2, corresponding to a 19.29\% relative RMSE reduction (RMSE Gain) caused by leakage. Similarly, at lag = 3, RMSE dropped from 2.8925 (clean) to 2.2987 (leaky), representing a 20.51\% apparent RMSE Gain (Table~\ref{tab:LeakageValidationExpanded}, Table~\ref{tab:rmse_drift_sensitivity}). These results indicate that leakage leads to overly optimistic performance estimates, particularly for longer forecasting horizons.

%%%%%%%%%%% figure Min RMSE %%%%%%%%%%%%%%%%%%%
\begin{figure}[H]
  \centering
  \graphicspath{{./images/}}
  \includegraphics[width=0.9\textwidth]{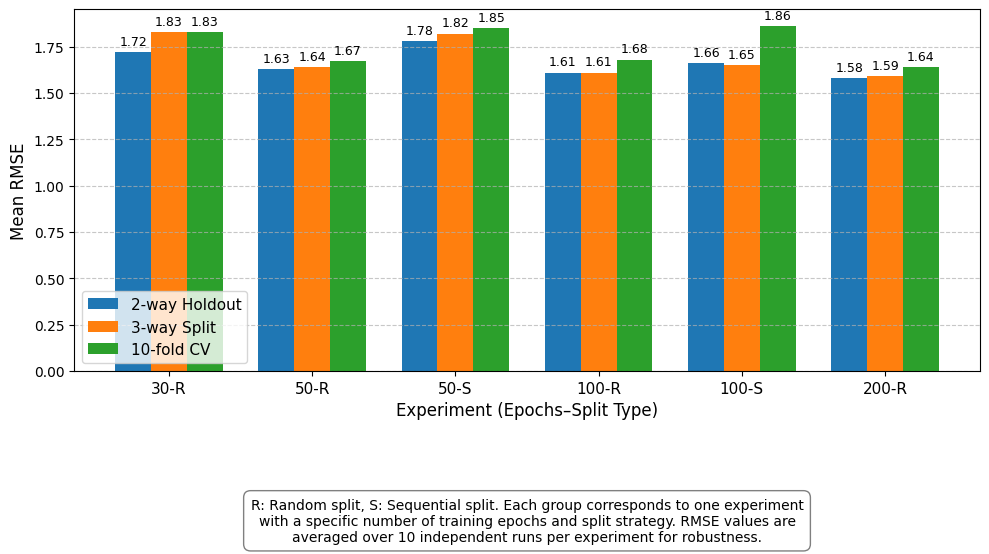}
  \caption{RMSE across different epochs with data leakage.} %
  \label{fig:baseline_rmse}
\end{figure}
%%%%%%%%%%%%%%%%%%%%%%%%%%%%%%%%%%%%%%

%%%%%%%%%%%%%%%%%%%%%%%%%%%%%%%%%%%%%%%%%%%%%%%%%%%%%%
%=================================

    \item \textbf{Window Size:} Smaller windows (e.g., size 3) showed an RMSE decrease from 1.7308 (clean) to 1.6476 (leaky) under 10-fold CV (4.81\% relative reduction), reflecting increased leakage sensitivity. Larger windows (e.g., size 7) also exhibited leakage-induced RMSE reduction from 1.7607 to 1.6347 (7.17\% gain), but overall model stability improved with larger temporal contexts (Table~\ref{tab:LeakageValidationExpanded}, Figure~\ref{Sequence}).
\end{itemize}

\begin{figure}[H]
  \centering
  \includegraphics[width=0.6\textwidth]{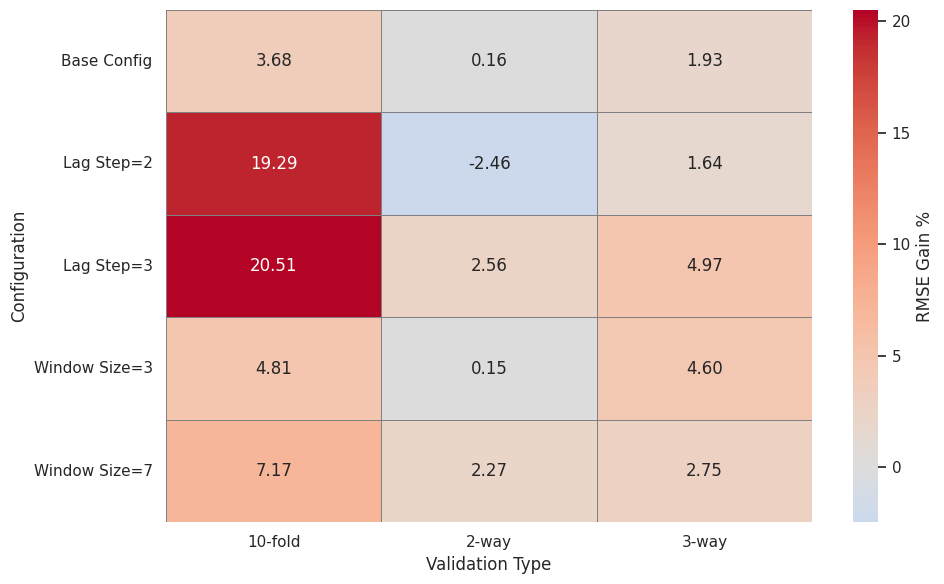}
  \caption{RMSE Gain Distributions Under Leaky Conditions}
  \label{3val}
\end{figure}

These observations confirm that inadequate temporal context (small windows) or longer forecasting horizons (higher lag steps) exacerbate vulnerability to leakage-induced bias in evaluation pipelines. When sequences are generated before data splitting (leaky setup), these factors distort model assessment by introducing data leakage. Conversely, under clean setups with sequences generated post-split, RMSE values were consistently more stable and the relative RMSE reduction due to leakage was minimal. For example, the base configuration using 2-way split showed only a 0.16\% RMSE difference between clean (1.6596) and leaky (1.6569) setups, whereas 10-fold CV showed a larger 3.68\% difference, highlighting the impact of validation strategy on leakage sensitivity (Table~\ref{tab:rmse_drift_sensitivity}). Figure~\ref{Sequence} further illustrates these contrasts, with 10-fold cross-validation showing the greatest RMSE gains and variability under leakage, especially at higher lag steps, while 2-way and 3-way splits demonstrate tighter confidence intervals and reduced leakage sensitivity (Figure~\ref{3val}). These findings emphasize the importance of proper dataset partitioning and architectural choices to mitigate evaluation bias caused by temporal leakage.

To assess the performance distortion introduced by data leakage, Table~\ref{tab:rmse_drift_sensitivity} reports RMSE values and relative RMSE Gain percentages comparing clean and leaky conditions for Experiment 5 across validation techniques. The most pronounced leakage effect was observed under 10-fold cross-validation with lag = 2, where RMSE decreased from 2.0868 (clean) to 1.6841 (leaky), corresponding to a substantial \textbf{19.29\% apparent RMSE Gain}. This highlights the susceptibility of fold-based validation to temporal leakage when sequence windows are generated prior to data splitting. In contrast, both 2-way and 3-way validation strategies showed greater robustness, with RMSE Gain generally below 3\% across configurations. A minor negative gain (–2.46\%) was detected for 2-way validation at lag = 2, likely reflecting stochastic noise rather than genuine improvement. These findings emphasize the utility of RMSE Gain for leakage sensitivity and underscore the importance of employing temporally consistent validation schemes in time series forecasting to avoid biased performance estimates.

%-------------------------------------------------------%

\begin{table}[htbp]
\centering
\caption{Performance of Experiment 5 and Its Variants With vs. Without Data Leakage Across Validation Techniques}
\label{tab:LeakageValidationExpanded}
\scriptsize
\setlength{\tabcolsep}{5pt}
\renewcommand{\arraystretch}{1.15}
\begin{tabular}{llllllc}
\toprule
\textbf{Test Type} & \textbf{Window} & \textbf{Lag} & \textbf{Leak} & \textbf{Val} & \textbf{Mean RMSE (95\% CI)} & \textbf{Std} \\
\midrule

Base Config & 10 & 1 & \checkmark & 2-way   & 1.6569 (1.6348, 1.6889) & 0.0447 \\
            &    &   & \checkmark & 3-way   & 1.6522 (1.5301, 1.7634) & 0.0595 \\
            &    &   & \checkmark & 10-fold & 1.8646 (1.8401, 1.8837) & 0.0305 \\
            &    &   & \xmark     & 2-way   & 1.6596 (1.6470, 1.6722) & 0.0176 \\
            &    &   & \xmark     & 3-way   & 1.6848 (1.6666, 1.6894) & 0.0159 \\
            &    &   & \xmark     & 10-fold & 1.9359 (1.9209, 1.9905) & 0.0487 \\
\midrule
Window Size & 7  & 1 & \checkmark & 2-way   & 1.6487 (1.6340, 1.6634) & 0.0205 \\
            &    &   & \checkmark & 3-way   & 1.6710 (1.6425, 1.6796) & 0.0259 \\
            &    &   & \checkmark & 10-fold & 1.6347 (1.6269, 1.6426) & 0.0109 \\
            &    &   & \xmark     & 2-way   & 1.6870 (1.6598, 1.7141) & 0.0379 \\
            &    &   & \xmark     & 3-way   & 1.7182 (1.6536, 1.7903) & 0.0955 \\
            &    &   & \xmark     & 10-fold & 1.7607 (1.7484, 1.7730) & 0.0172 \\
\midrule
Window Size & 3  & 1 & \checkmark & 2-way   & 1.6493 (1.6381, 1.6604) & 0.0155 \\
            &    &   & \checkmark & 3-way   & 1.6516 (1.6325, 1.6706) & 0.0266 \\
            &    &   & \checkmark & 10-fold & 1.6476 (1.6363, 1.6589) & 0.0158 \\
            &    &   & \xmark     & 2-way   & 1.6518 (1.6374, 1.6663) & 0.0202 \\
            &    &   & \xmark     & 3-way   & 1.7312 (1.6922, 1.7540) & 0.0432 \\
            &    &   & \xmark     & 10-fold & 1.7308 (1.7221, 1.7396) & 0.0122 \\
\midrule
Lag Step    & 10 & 2 & \checkmark & 2-way   & 1.7964 (1.7776, 1.8152) & 0.0262 \\
            &    &   & \checkmark & 3-way   & 1.8492 (1.8322, 1.8662) & 0.0237 \\
            &    &   & \checkmark & 10-fold & 1.6841 (1.6752, 1.6930) & 0.0123 \\
            &    &   & \xmark     & 2-way   & 1.7533 (1.7213, 1.7853) & 0.0448 \\
            &    &   & \xmark     & 3-way   & 1.8800 (1.7729, 1.8892) & 0.0813 \\
            &    &   & \xmark     & 10-fold & 2.0868 (2.0100, 2.6868) & 0.4728 \\
\midrule
Lag Step    & 10 & 3 & \checkmark & 2-way   & 1.9804 (1.9534, 1.9975) & 0.0377 \\
            &    &   & \checkmark & 3-way   & 2.4802 (2.4631, 2.4973) & 0.0239 \\
            &    &   & \checkmark & 10-fold & 2.2987 (2.2775, 2.3199) & 0.2960 \\
            &    &   & \xmark     & 2-way   & 2.0325 (2.0037, 2.0613) & 0.0403 \\
            &    &   & \xmark     & 3-way   & 2.6100 (2.1253, 2.8597) & 0.5136 \\
            &    &   & \xmark     & 10-fold & 2.8925 (2.8923, 2.9989) & 0.0746 \\
\bottomrule
\end{tabular}
\caption*{\scriptsize \xmark: Data split before sequence creation (no leakage); \checkmark: Sequence created before data split (leakage). Val = Validation Technique.}
\end{table}
%--------------------------------%

Beyond validation strategy, architectural parameters such as lag step and window size had a significant impact on model performance and leakage sensitivity. Larger lag steps consistently resulted in higher RMSE values and amplified RMSE Gain, highlighting the increased difficulty of long-horizon forecasting under leaky evaluation conditions. The effect of window size was nonlinear: smaller windows (e.g., size 3) produced less stable predictions with greater leakage-induced RMSE reduction, whereas larger windows (e.g., size 10) enhanced performance and reduced the RMSE disparity between clean and leaky setups. These results indicate that providing models with a broader temporal context mitigates vulnerability to leakage and improves robustness across both clean and leaky evaluation protocols. Moreover, the RMSE Gain metric emerges as a practical and sensitive indicator for detecting leakage-prone configurations and uncovering evaluation bias in time series forecasting pipelines.

%%%%%%%%%%% Methodology Overview %%%%%
%%%%%%%%%%%%%%%%%%%%%%%%%%%%%%%%%%%%%%

%%%%%%%%%%% Methodology Overview %%%%%
%%%%%%%%%%%%%%%%%%%%%%%%%%%%%%%%%%%%%%

%=================================

\begin{table}[htbp]
\centering
\caption{RMSE Gain and Leakage Sensitivity Ranking}
\label{tab:rmse_drift_sensitivity}
\scriptsize
\setlength{\tabcolsep}{5pt}
\renewcommand{\arraystretch}{1.2}
\begin{tabular}{lllllllc}
\toprule
\textbf{Test} & \textbf{Win} & \textbf{Lag} & \textbf{Val} & \textbf{Clean} & \textbf{Leaky} & \textbf{Gain \%} & \textbf{Leakage Rank} \\
\midrule
Base Config   & 10 & 1 & 2-way    & 1.6596 & 1.6569 & ↑ 0.16  & 1 \\
              &    &   & 3-way    & 1.6848 & 1.6522 & ↑ 1.93  & 2 \\
              &    &   & 10-fold  & 1.9359 & 1.8646 & ↑ 3.68  & 3 \\
\midrule
Window Size   & 3  & 1 & 2-way    & 1.6518 & 1.6493 & ↑ 0.15  & 1 \\
              &    &   & 3-way    & 1.7312 & 1.6516 & ↑ 4.60  & 3 \\
              &    &   & 10-fold  & 1.7308 & 1.6476 & ↑ 4.81  & 2 \\
\midrule
Window Size   & 7  & 1 & 2-way    & 1.6870 & 1.6487 & ↑ 2.27  & 1 \\
              &    &   & 3-way    & 1.7182 & 1.6710 & ↑ 2.75  & 2 \\
              &    &   & 10-fold  & 1.7607 & 1.6347 & ↑ 7.17  & 3 \\
\midrule
Lag Step      & 10 & 2 & 2-way    & 1.7533 & 1.7964 & ↓ 2.46  & 1 \\
              &    &   & 3-way    & 1.8800 & 1.8492 & ↑ 1.64  & 2 \\
              &    &   & 10-fold  & 2.0868 & 1.6841 & ↑ 19.29 & 3 \\
\midrule
Lag Step      & 10 & 3 & 2-way    & 2.0325 & 1.9804 & ↑ 2.56  & 1 \\
              &    &   & 3-way    & 2.6100 & 2.4802 & ↑ 4.97  & 2 \\
              &    &   & 10-fold  & 2.8925 & 2.2987 & ↑ 20.51 & 3 \\
\bottomrule
\end{tabular}

\caption*{\scriptsize
\textbf{Win} = Window size; \textbf{Val} = Validation technique;  
\textbf{Gain \%} = $\frac{\mathrm{Clean} - \mathrm{Leaky}}{\mathrm{Clean}} \times 100$.  
\textbf{Leakage Rank} orders configurations by absolute RMSE change magnitude (1 = lowest, 3 = highest leakage sensitivity).  
Positive drift (↑) indicates improved performance due to leakage; negative drift (↓) may reflect stochastic variability or noise.
}
\end{table}

Findings from Phases I--III emphasize the critical importance of validation design in mitigating data leakage within time series forecasting. Specifically, the application of shuffled 10-fold cross-validation without enforcing temporal boundaries—such as non-overlapping splits or buffer zones—can lead to significant overestimation of model performance caused by future information leakage~\cite{f1,f2}. To counteract this, several best practices are advocated. Foremost, \textit{post-split sequence generation} must be applied to maintain strict temporal separation between training and testing data. When fold overlap cannot be avoided, incorporating a \textit{buffer zone} between folds effectively reduces leakage risk. Furthermore, \textit{2-way} and \textit{3-way validation} methods demonstrate superior robustness to temporal contamination, especially when implemented with sequential data partitioning. Lastly, reporting \textit{RMSE Gain} alongside \textit{confidence intervals} provides a valuable means to detect evaluation bias and improves the interpretability and reliability of forecasting assessments.

%As summarized in Table~\ref{tab:rmse_drift_sensitivity}, 2-way and 3-way validation strategies consistently exhibit lower RMSE Drift values and demonstrate greater robustness to leakage, while 10-fold cross-validation shows significantly higher drift and thus greater susceptibility to leakage—particularly at larger lag steps.
.

\section{Future Work and Research Directions}
\label{sec:future_work}

This study establishes a systematic framework for analyzing data leakage in LSTM-based time series forecasting, yet several avenues remain to broaden its applicability and generalization. The following directions are proposed to extend the scope of this research:

\paragraph{Controlled Leakage as a Diagnostic Mechanism.} While this work treats leakage as a methodological flaw, future studies may explore its controlled use as a stress-testing mechanism. In safety-critical applications, such as healthcare or autonomous systems, intentionally simulating configuration-aware leakage—e.g., limited temporal overlap in blocked validation—could provide insight into model behavior under operational uncertainty and serve as a diagnostic for robustness evaluation.

\paragraph{Architectural Generalization Beyond LSTM.} The current analysis focuses on LSTM models. Future investigations should assess the leakage sensitivity of contemporary architectures, including Temporal Convolutional Networks (TCNs), Transformer-based models (e.g., Informer, Autoformer), and large language models (LLMs) adapted for temporal prediction. As these architectures are increasingly adopted in cross-domain contexts, their evaluation must conform to strict temporal protocols to ensure reliability and fairness.

\paragraph{Development of Leakage-Resistant Validation Protocols.} The findings underscore the limitations of conventional cross-validation techniques in sequential settings. Future work should focus on developing and benchmarking temporally consistent evaluation schemes, such as forward-chaining, blocked k-fold, and nested time series cross-validation. Empirical comparisons across forecast horizons and temporal granularities will be necessary to identify best practices.

\paragraph{Dataset Diversity and Benchmark Generalization.} Generalizability beyond the Climate dataset used in this study requires validation across a broader range of domains. Expanding to datasets in energy, finance, and clinical monitoring will help determine whether leakage effects are architecture-specific or manifest universally across time-dependent tasks.

\paragraph{Regularization under Leakage Conditions.} Further investigation is warranted into how data augmentation and noise injection techniques (e.g., jittering, masking, mixup) influence model behavior under clean versus leaky conditions. Understanding this interaction may inform the development of robust regularization strategies and leakage-aware evaluation frameworks.

\paragraph{Automated Detection of Temporal Leakage.} As ML pipelines become increasingly automated, the development of tools for static or dynamic verification of temporal integrity is essential. Such tools could flag violations in sequence construction or partitioning logic, supporting reproducible and transparent model validation.

%Taken together, these research directions aim to strengthen the empirical foundation for leakage-aware evaluation and promote the development of temporally robust forecasting systems. By advancing diagnostic tools, benchmarking standards, and architecture-agnostic protocols, future work can enhance both methodological reliability and deployment readiness in sequential learning.

%%%%%%%%%%%%%%

%---------------------------------------------%

\section{Conclusion}
\label{sec:conclusion}

This study presented a systematic evaluation of data leakage in LSTM-based time series forecasting, emphasizing the critical role of validation design and sequence handling in ensuring trustworthy performance estimation. Through a controlled experimental framework, we demonstrated that pre-split sequence generation—commonly overlooked in practice—can introduce substantial temporal leakage and bias evaluation results. To measure this, \textit{RMSE Gain} is used to show the relative change in performance between clean and leaky setups. Empirical results revealed that 10-fold cross-validation, when applied without temporal safeguards, is particularly vulnerable to leakage, with RMSE Gain reaching over 20.5\% in certain settings. By contrast, 2-way and 3-way validation techniques exhibited significantly lower susceptibility, with 2-way validation demonstrating the greatest robustness across varying input window sizes and forecast horizons. These findings underscore the importance of leakage-aware evaluation pipelines in time series modeling. Sequence generation must be performed after data partitioning to maintain temporal causality. Additionally, reporting RMSE Gain alongside conventional metrics helps identify evaluation artifacts. The framework introduced in this work offers a reproducible methodology for detecting and mitigating leakage effects and serves as a foundation for more reliable benchmarking in temporal forecasting. These contributions support the development of robust, transparent, and generalizable evaluation practices for sequential deep learning models in both research and deployment contexts.

%%%%%%%%%%%%%%%%%%%%%%%%%%%%%%%%%%%%%%%%%%%%%%%%%%%%%%%%%%%%%%%%%%%%

%---------------------------------------------%
\begin{credits}
\subsubsection{\ackname} The authors would like to acknowledge the support received from the Saudi Data and AI Authority (SDAIA) and King Fahd University of Petroleum \& Minerals (KFUPM) under the SDAIA-KFUPM Joint Research Center for Artificial Intelligence Grant JRC-AI-RFP-20.

\subsubsection{\discintname}
 The authors have no competing interests to declare that are relevant to the content of this article.
\end{credits}
% ---- Bibliography ----
%
% BibTeX users should specify bibliography style 'splncs04'.
% References will then be sorted and formatted in the correct style.
%
% \bibliographystyle{splncs04}
% \bibliography{mybibliography}
\bibliographystyle{splncs04}
\bibliography{Ref}
%
%\begin{thebibliography}{8}
%\bibitem{ref_article1}
%Author, F.: Article title. Journal \textbf{2}(5), 99--110 (2016)

%\bibitem{ref_lncs1}
%Author, F., Author, S.: Title of a proceedings paper. In: Editor,
%F., Editor, S. (eds.) CONFERENCE 2016, LNCS, vol. 9999, pp. 1--13.
%Springer, Heidelberg (2016). \doi{10.10007/1234567890}

%\bibitem{ref_book1}
%Author, F., Author, S., Author, T.: Book title. 2nd edn. Publisher,
%Location (1999)

%\bibitem{ref_proc1}
%Author, A.-B.: Contribution title. In: 9th International Proceedings
%on Proceedings, pp. 1--2. Publisher, Location (2010)

%\bibitem{ref_url1}
%LNCS Homepage, \url{http://www.springer.com/lncs}, last accessed 2023/10/25
%\end{thebibliography}
\end{document}